\documentclass[letterpaper, 10 pt, conference]{ieeeconf}  

\IEEEoverridecommandlockouts                              

\usepackage{graphicx} 
\usepackage{tabularx}
\usepackage{comment}
\usepackage{xspace}
\usepackage{balance}
\usepackage{amsmath}
\usepackage{hyperref} 

\newcommand{\system}{HEXAR\xspace}

\title{\LARGE \bf
HEXAR: a Hierarchical Explainability Architecture for Robots
}

\author{Tamlin Love$^{1*}, $Ferran Gebellí$^{2*}, $Pradip Pramanick$^{3*}, $\\Antonio Andriella$^{1}, $Guillem Alenyà$^{1}, $Anaís Garrell$^{1}, $Raquel Ros$^{4}, $Silvia Rossi$^{3}$
\thanks{$^{*}$These authors contributed equally}%
\thanks{$^{1}$Institut de Robòtica i Informàtica Industrial (CSIC-UPC), Llorens i Artigas 4-6, 08028, Barcelona, Spain}%
\thanks{$^{2}$PAL Robotics, Pujades, 77-79, 7-7, 08005 Barcelona, Spain}%
\thanks{$^{3}$University of Naples Federico II, Naples, Italy}%
\thanks{$^{4}$Artificial Intelligence Research Institute (IIIA-CSIC), Barcelona, Spain}%
}

\begin{document}

\maketitle
\thispagestyle{empty}
\pagestyle{empty}

\begin{abstract}

As robotic systems become increasingly complex, the need for explainable decision-making becomes critical. Existing explainability approaches in robotics typically either focus on individual modules, which can be difficult to query from the perspective of high-level behaviour, or employ monolithic approaches, which do not exploit the modularity of robotic architectures. We present \system (Hierarchical EXplainability Architecture for Robots), a novel framework that provides
a plug-in, hierarchical approach to generate explanations about robotic systems. \system consists of specialised component explainers using diverse explanation techniques (e.g., LLM-based reasoning, causal models, feature importance, etc) tailored to specific robot modules,
orchestrated by an explainer selector that chooses the most appropriate one for a given query. We implement and evaluate \system on a TIAGo robot performing assistive tasks in a home environment, comparing it against end-to-end and aggregated baseline approaches across 180 scenario-query variations. We observe that \system significantly outperforms baselines in root cause identification, incorrect information exclusion, and runtime, offering a promising direction for transparent autonomous systems.

\end{abstract}

\section{Introduction}
\label{sec:introduction}

Robotic software systems are inherently complex, typically comprising architectures with numerous modules that accomplish diverse capabilities and employ various interfaces \cite{macenski2022robot}.
Despite the emerging end-to-end learning approaches \cite{ma2024survey, kim2024openvla, team2025gemini}, where a single black-box module processes sensor data to produce near-final actuator signals, modular architectures remain essential for providing structured, aggregated and meaningful information channels. These internal signals serve as intermediate and efficient means to transmit information, whilst also functioning as internal data representations that facilitate human understanding.
For instance, in object navigation tasks, while pure end-to-end approaches may perform well in simulation, modular approaches perform better in real-world scenarios~\cite{gervet2023navigating}, with the additional benefit of informative intermediate signals such as the planned path.\looseness=-1

The ability to explain the decisions and behaviours related to these internal processes has been acknowledged as a key factor for improving human understanding of eXplainable Artificial Intelligence (XAI) systems \cite{miller2019explanation} and autonomous robots \cite{anjomshoae2019explainable}. However, existing explainability approaches in robotics focus on individual modules such as addressee selection \cite{bevckova2025multi}, navigation \cite{halilovic2023visuo}, manipulation \cite{duan2024aha}, motion planning~\cite{brandao2021towards}, or task planning \cite{krarup2019model}. These approaches typically evaluate isolated scenarios wherein only the module in question is utilised. In the context of a general-purpose robot, end-users lack knowledge of internal robot architectures unless explicitly informed \cite{hindemith2025improving} and seek explanations for high-level behaviours, which potentially involve multiple components. Consequently, people tend to pose general ``Why'' questions~\cite{wachowiak2024people} (e.g., ``Why did the robot not go to the kitchen''), rather than directing their explanation-seeking queries to specific components (e.g., ``Why did pose estimation fail?'').

\begin{figure} [t]
\centering
\includegraphics[width=0.48\textwidth]{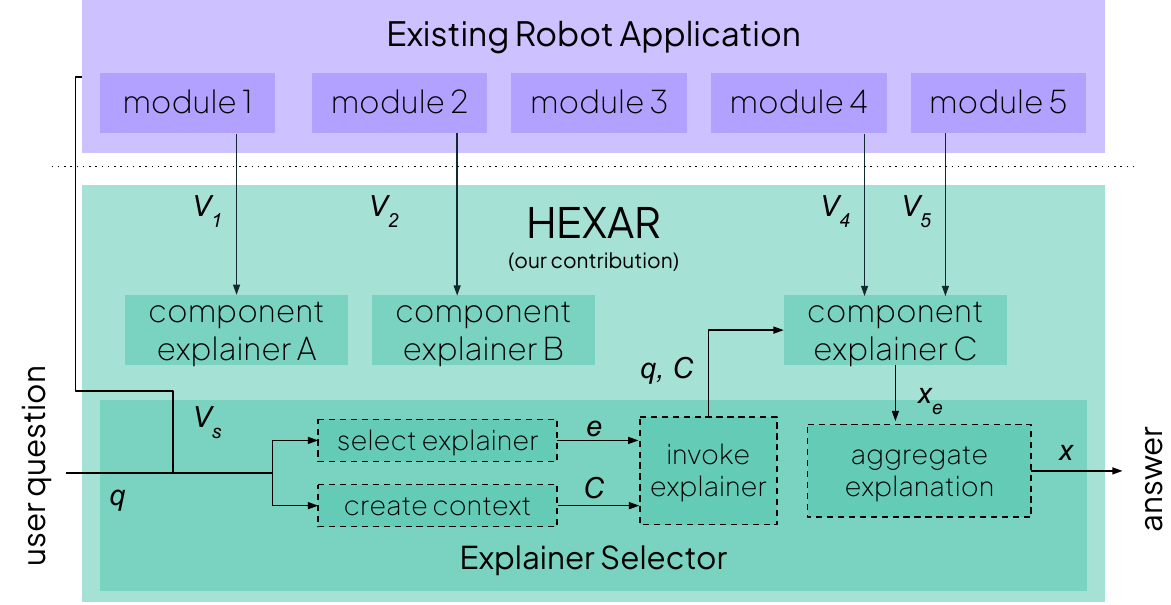}
\caption{Our novel framework, HEXAR, in which different specialised component explainers are orchestrated by an explainer selector, which selects an appropriate component explainer to answer a given query.}
\label{fig:architecture_overview}
\end{figure}

There exist some system-wide explainability approaches that employ monolithic methods whereby centralised Large Language Models (LLMs) exploit robot logs and other diagnostic information~\cite{liu2023reflect, wang2025can, sobrin2024explaining}, or that assume a single behaviour tree representation \cite{lemasurier2024templated}. However, it has been argued that explainability in robotics should aim for modular architectures and hybrid models \cite{winikoff2024towards, adebayo2024explainable}.
Our contribution is \system, a hierarchical explainability framework (Fig. \ref{fig:architecture_overview}). This framework allows a system to utilise different specialised explainers, each corresponding to one or more application components. The system selects the most appropriate explainer based on the user query and other contextual factors. This approach motivates our research question: \textit{Is a hierarchical explanation system, composed of specialised explainer modules orchestrated by a selector module, better at accurately explaining complex and modular robotic systems than equivalent monolithic systems?}\looseness=-1

To answer this question, we implement our approach (described in Sec. \ref{sec:framework}) in a home assistant robot use case (Sec. \ref{sec:implementation}) and evaluate the generated explanations across a range of scenarios (Sec. \ref{sec:experiments})\footnote{The annotated datasets, implementation and reproducible results are in \url{https://anonymous.4open.science/r/HEXAR-CEF2}}. We provide evidence that our approach produces more accurate explanations in less time than two baselines (Sec. \ref{sec:discussion}), laying a foundation for explainable autonomous robot architectures.

\section{Related Work}
\label{sec:literature}
While the automatic generation of robot explanations has been identified as an important research area~\cite{anjomshoae2019explainable}, many existing approaches target specific robot modules (software components that target specific functionalities) rather than considering entire systems. For example, in~\cite{bevckova2025multi}, robot addressee selection is explained using attention maps, feature importance and decision confidence. In \cite{halilovic2023visuo}, robot navigation is explained using a mix of ontologies, qualitative spatial reasoning \cite{freksa1991qualitative}, and LIME \cite{ribeiro2016should}, which is a widely used method for producing feature-importance explanations for black-box models. The work in \cite{duan2024aha} presents an architecture for generating explanations about manipulation tasks employing a Visual Language Model (VLM). In another work \cite{brandao2021towards}, motion planning is explained using a templated approach that exploits planning constraints and algorithm parameters. To explain general task plans, the method presented in \cite{krarup2019model} produces contrastive explanations by generating a contrastive PDDL plan. From these examples and the wider literature~\cite{sobrin2025generating}, it is clear that 
many approaches (which are usually module-specific) employ very different explanation generation techniques, with some being more appropriate than others for particular functionalities. One important feature of our proposed framework is that it allows the combination of different explanation techniques.\looseness=-1

Some other works have proposed broader explanation modules that can explain a full robotic system, usually employing LLMs. In \cite{sobrin2024explaining}, all the robot's logs are collected by an LLM, which can then answer user queries about the system. Although this approach is presented as generic for any robot using ROS2 \cite{macenski2022robot}, it is validated only in a navigation scenario. The REFLECT approach \cite{liu2023reflect} goes further and constructs a robot event summary based on the robot plan, a scene graph built from RGBD images, and audio data. This event summary is then used to verify success for each plan subgoal, and triggers two different LLM prompts depending on whether a subgoal failed (execution analysis) or not (planning analysis). Similarly, the RONAR \cite{wang2025can} framework uses the robot plan, an RGBD-based scene graph, and base/joint states to generate narration summaries that can be used to generate explanations about failures. The RACCOON framework \cite{bustamante2025raccoon} generates explanations of system-level behaviour by first selecting relevant robot modules, whose information is passed to an LLM for explanation generation, taking advantage of the application modularity, although all the modules are explained by the same centralised LLM. In another approach \cite{lemasurier2024templated}, the whole robot decision process is assumed to be embedded in a single behaviour tree. Then, given the predefined possible sequences of actions represented by the behaviour tree, either templated or LLM-generated explanations can provide questions to general user queries.\looseness=-1


These monolithic approaches do provide explanations to general questions about a robot's behaviour, but do not allow for specialised explainers that target specific robot functionalities and that might employ tailored techniques for the explanation generation.  
There have been a few initial propositions that go in the direction of such a modular explainability architecture, though they do not fully implement it. 
The theoretical work in \cite{winikoff2024towards} advocates for such an architecture while presenting several integration issues, including how to orchestrate different explainers, how to connect those explainers to the application components, or how to store the relevant data over time.
The work in \cite{stange2022self}
defines and validates an architecture for explainable behaviour generation. Within the requirements, it mentions
component-level inspectability and interpretable inter-component communication interfaces. However, the presented architecture remains a unique centralised module that combines information from the decision-making and episodic memory. In COPAL \cite{joublin2024copal}, a system architecture orchestrates 3 cognitive levels for reasoning, planning, and motion generation, respectively. Each level is implemented as an LLM agent that interacts with the others. Nevertheless, this hierarchical approach is never applied to explainability, which is mentioned as a feature for future work. Finally, SNAPE \cite{robrecht2023snape} formalises the explanation generation process as inherently hierarchical. Nonetheless, the purpose of this hierarchy is to have local Markov Decision Processes (MDPs) provide online explanations adapted to changes in the interaction, while the source explainability information is pre-computed in a global explanation plan.

In this work, we present and validate a framework that takes advantage of a robotic system's modularity to implement a hierarchical explanation generation architecture that is able to answer general questions about a robot's behaviour and decision-making process by selecting the most appropriate specialised component explainers.\looseness=-1

\section{Hierarchical Explanation Framework}
\label{sec:framework}

We propose a novel hierarchical system for explaining a robot's behaviours and decision-making in response to a user query. The system, which we name \system (\emph{\underline{H}ierarchical \underline{EX}plainability \underline{A}rchitecture for \underline{R}obots}), is composed of a set of specialised \textbf{component explainers} $\mathcal{E}$, which can provide explanations for one or more robot modules (encompassing both high- and low-level modules), and an \textbf{explainer selector} $s$, which selects a subset $\mathcal{E}_s \subseteq \mathcal{E}$ of component explainers that should answer a given query $q$. We assume that the robot system emits a sequence of events $\mathcal{V}$. Each component explainer $e$ independently observes a subset $\mathcal{V}_e \subseteq \mathcal{V}$ of events, as does the explainer selector, which observes $\mathcal{V}_s \subseteq \mathcal{V}$. In practice, this observation could be implemented in a publisher-subscriber architecture, such as component explainer nodes listening to topics in ROS. An overview of the presented framework can be seen in Fig. \ref{fig:architecture_overview}.\looseness=-1

\textbf{Component explainers} are responsible for generating explanations targeting either execution-oriented skills or actions (e.g., navigation, manipulation, human interaction) or higher-level modules of the robot architecture (e.g., task planning, world modelling). We can represent a component explainer as a tuple $e = \langle f_e, \mathcal{V}_e \rangle$, where $f_e(q,C)$ is a function that takes in a user query $q$ and a context vector $C$ (containing, for example, task information, time windows, etc.) and returns a natural language explanation $x_e$. We note that the interface by which $f_e$ is called is agnostic to its underlying implementation, which could use a diverse set of explanation generation techniques (e.g., counterfactual, feature importance, templated, etc.). The implementation of $f_e$ may involve the invocation of other processes or components. For example, if an interaction module invokes a grasping module to perform object handover, the component explainer assigned to the interaction module could invoke the one assigned to the grasping module, if required.

Necessarily, \system requires a mapping $M$ from the set of robot modules to $\mathcal{E}$. It may be that a single component explainer is responsible for multiple robot modules, perhaps because they can be explained in a similar manner or rely on the same data. It is also possible for multiple component explainers to be assigned to a single robot module, representing different approaches to explaining the module. If it is not required to explain a particular robot module, it is not necessary to implement a corresponding component explainer.\looseness=-1


Upon receiving a user query $q$, the \textbf{explainer selector} $s$ performs three functions in sequence to produce a natural language explanation $x$. Firstly, it is responsible for selecting an appropriate set $\mathcal{E}_s \subseteq \mathcal{E}$ of component explainers, using some selection function $select(q,\mathcal{V}_s)$. The exact selection logic itself is use case dependent. It may follow a heuristic (e.g., selecting component explainers for modules that have failed) or be determined by a classifier (e.g., an LLM). If multiple component explainers are available for a single robot module, this selection logic may choose between them using information in $\mathcal{V}_s$ (such as different tasks, user types, queries or interaction states).

Additionally, the explainer selector provides a context vector $C$, computed from the query $q$ and observed events $\mathcal{V}_s$, which is passed to the selected component explainers. The context vector may include information such as the relevant task to explain, the time window in which a module is executed, etc.

Finally, after executing $f_e, \quad \forall e \in \mathcal{E}_s$, the explainer selector must aggregate the set of explanations $\{x_e |e\in\mathcal{E}_s\}$ into a single explanation $x$ if $|\mathcal{E}_s|>1$. The aggregation method may be implemented in a number of ways, for example, using an LLM to summarise the explanation set.

As a complete framework, \system is designed as an add-on to any existing modular robot architecture, making use of already existing information flows ($\mathcal{V}$), such as diagnostics logs or interfaces between application modules. The main advantage of this plug-in approach is that the application implementation and architecture remain the same, without the need for modification. Of course, the quality of explanations may be limited by the quality and level of detail of the information made available by the robot. Depending on the use case, greater explanation coverage may be achieved by augmenting the logging and other diagnostic interfaces of the robot modules. Here, we reinforce the argument for continuously logging key events from sensing, actuation, and decision-making subsystems~\cite{winikoff2024towards,winfield2017case}, given that generic data logging can be beneficial to explain unforeseen factors beyond what is anticipated for each module.\looseness=-1


To design the explainability add-on and decide the specific architecture and required component explainers, several methodologies can be used to identify the user explainability needs and necessary internal information sources, such as co-design \cite{gebelli2024co} or explainability-by-design \cite{huynh2022explainability}.\looseness=-1

\section{Implementation}\label{sec:implementation}
To evaluate the proposed framework, we implement it on a physical robotic platform within a home-assistance use case.

\subsection{Use Case}\label{sec:implementation_use_case} 
\begin{figure} [t]
\centering
\includegraphics[width=0.46\textwidth]{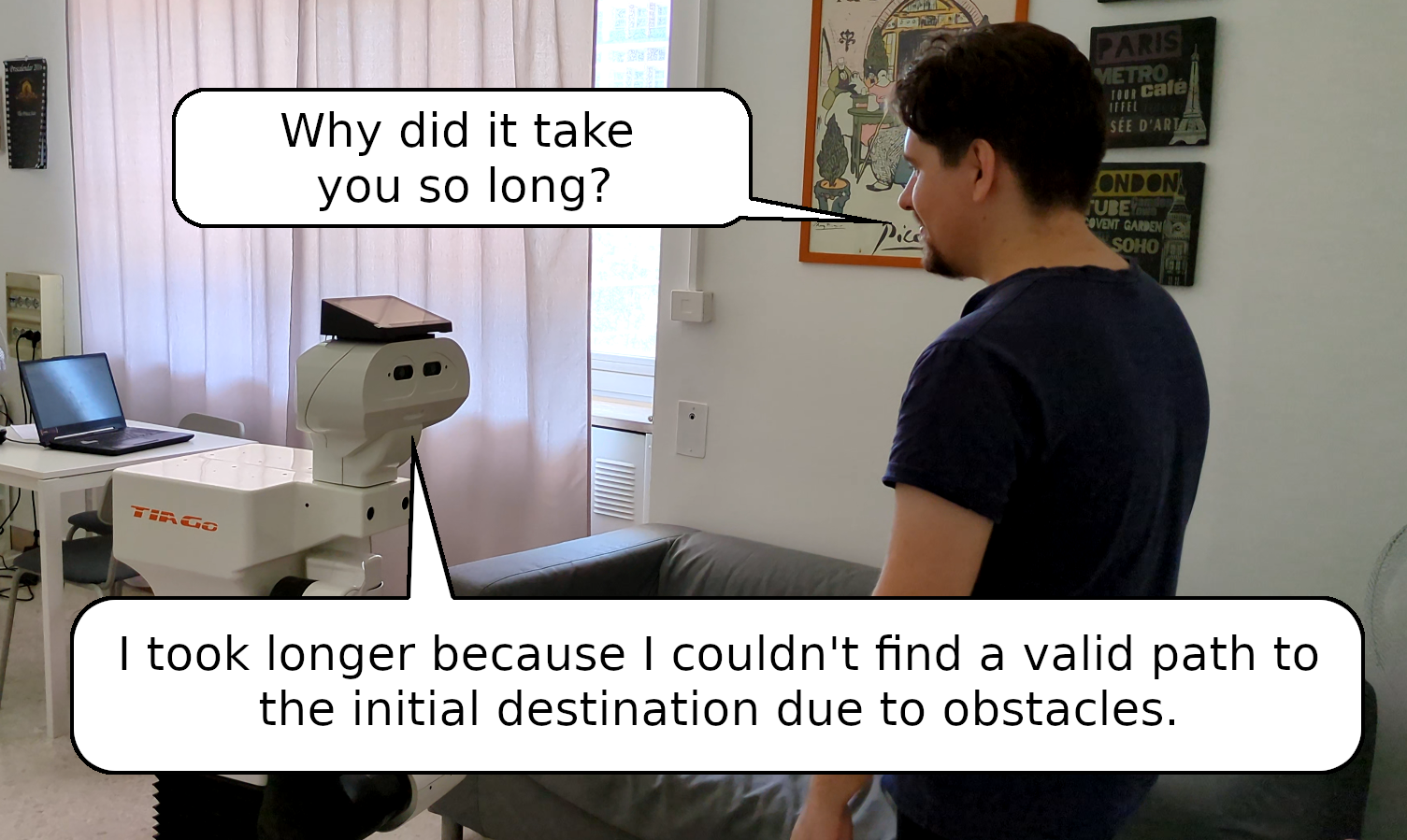}
\caption{Use case of a Tiago robot that assists in a home environment.}
\label{fig:tiago_photo}
\end{figure}

We use a TIAGo robot, as depicted in Fig. \ref{fig:tiago_photo}, which we equip with 4 skills and a task planner, each implementing a robot module. The skills we use in this evaluation are — navigating to different rooms (\textit{navigation}), speaking (\textit{text to speech}), asking humans for help to complete simple objectives (\textit{ask human for help}), and recommending pizza recipes based on available ingredients (\textit{pizza recommender}). Some skills are themselves complex and hierarchical. For example, the \textit{ask human for help} skill, implemented using a finite state machine, encompasses detecting humans, approaching one, asking for help, and confirming that the human has helped, making use of the \textit{navigation} and \textit{text to speech} skills. Other skills are relatively simple, such as the \textit{pizza recommender}, which is implemented using a decision tree classifier trained on a small dataset of pizza recipes. To this system we add an ``\textit{explain}'' skill, which triggers the \system framework (see Sec. \ref{sec:framework_instantiation}).\looseness=-1 

Using these skills, the robot can perform several assistive tasks such as bringing objects from one room to another, delivering messages to other people, or holding conversations with users. Achieving these tasks is facilitated by a \textit{task planner} \cite{cooper2025demonstration}, which converts a user request to a sequence of skills.\looseness=-1

\subsection{Framework Instantiation}\label{sec:framework_instantiation}
\begin{figure} [tb]
\centering
\includegraphics[width=0.49\textwidth]{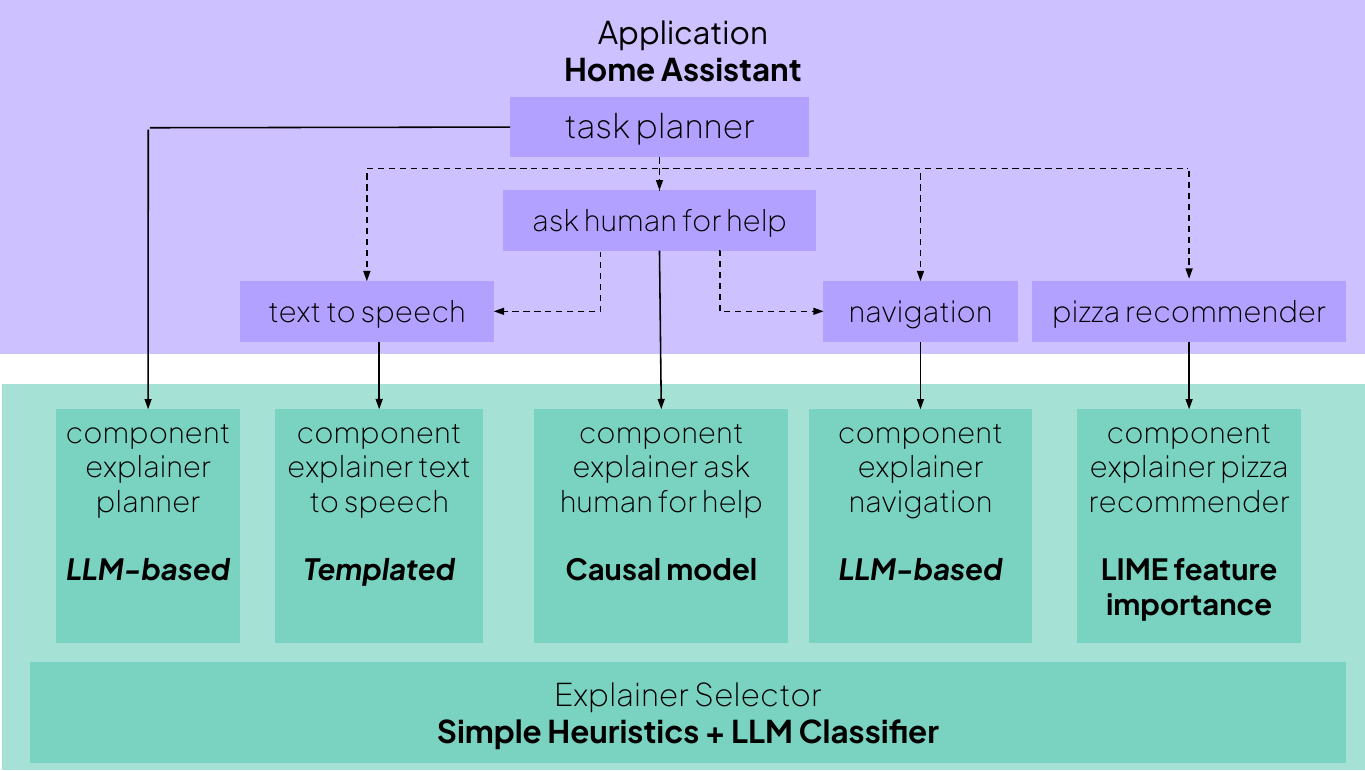}
\caption{The \system framework applied to the home assistant use case, showing the dependencies between modules. A component explainer is implemented for each robot module, employing a variety of explanation techniques.}
\label{fig:architecture_implementation}
\end{figure}

For each of the five robot modules discussed in Sec. \ref{sec:implementation_use_case}, we design and implement a component explainer that is tailored to answer queries about the specific skill. To illustrate that our framework can incorporate arbitrary component explainers, each of them uses a different explanation generation mechanism. Fig.  \ref{fig:architecture_implementation} provides an overview of the implemented architecture.

\looseness=-1

Firstly, the \textit{component explainer planner} is responsible for generating explanations for queries related to task planning. We utilise the broad ``common sense'' reasoning capabilities demonstrated by LLMs, similar to previous works on explanation generation for task planning~\cite{liu2023reflect,wang2025can}. In particular, we prompt an LLM to generate the explanation by providing a context that includes the human instruction, generated task plan, and errors in grounding it to a skill sequence (if any), status of each skill (e.g., \textit{waiting, failed}), and the user's explanation-seeking query. This open-ended prompting with the user query allows this component to address various types of plan issues (see Table~\ref{tab:dataset} for examples).

The component explainer for the \textit{text-to-speech} skill adopts a straightforward, template-based strategy that addresses a single failure condition, namely a skill timeout arising from excessively long utterances. In such instances, the explainer explicitly notifies the user of the timeout event.

Inspired by the work introduced in~\cite{sobrin2024explaining}, we implement the component explainer for the \textit{navigation} component with an LLM that receives mainly the logs from the ROS2 navigation Nav2 package~\cite{macenski2020marathon2}. Repetitive and irrelevant logs are discarded based on the known string patterns. The LLM is given an exhaustive list of examples, including possible failure and suboptimal situations that the logs can reveal, in order to provide explanations for user questions.

To implement the component explainer for the \textit{ask human for help} skill, we adapt the causal model approach of \cite{love2025temporal}, representing a family of approaches for contrastive, causal and counterfactual explanations. The method automatically builds a causal model of the skill execution and queries to generate counterfactual explanations of the form ``$Y$ occurred because $X=x$. If $X=x^*$, $Y^*$ would have occurred instead'' in response to the query ``Why $Y$ and not $Y^{*}$''. In failure cases, explanations provide changes that would result in success. These counterfactual templates are converted to a more natural explanation with an LLM. If no failure occurred, the model is consulted to determine if particular negative conditions were met, such as a high variance in the human detection, for which a templated explanation is given.\looseness=-1

The final component explainer, tailored to the \textit{pizza recommender} skill, makes use of LIME \cite{ribeiro2016should}, a widely-used explainer which ranks input features based on their relevance to a decision. In this case, LIME is used to determine the most relevant ingredient used to recommend a given type of pizza.\looseness=-1

To implement the explainer selector, we define $select(q,\mathcal{V}_s)$ as a two-stage process. Firstly, the explainer selector fetches the last available plan (in the form of a sequence of skills and their execution statuses) and checks if one of the skill executions failed or if the plan itself is invalid. If so, the corresponding component explainer is automatically selected (the component explainer planner in the case of invalid plans). Otherwise, if no failure or invalid plan is detected in the skill executions, the user's query is passed to an LLM, which selects the component explainer that corresponds best to the content of the query. Thus, this particular implementation selects only a single component explainer, negating the need for explanation aggregation.

The context vector $C$ consists of the task being explained, the sequence of skills and their return statuses, the plan validity, and the time window of the task. This context is provided to the selected component $e$ explainer when invoking $f_e$.\looseness=-1

In this section, we have presented one possible implementation of the general and flexible framework presented in Sec. \ref{sec:framework}. We emphasise that the concrete realisation of the framework can be highly tailored to the specific use case, domain constraints, user requirements and final implementation choices.\looseness=-1

\section{Experiments}
\label{sec:experiments}
In this section, we evaluate explanations generated by \system in the use case presented in Sec. \ref{sec:implementation} and compare them against baseline approaches. To demonstrate that \system can run locally, we have used \textit{phi4} \cite{abdin2024phi} (a model with $14B$ parameters that can run on consumer-grade GPUs) for all LLMs within the explanation add-on and application. We further use greedy decoding in the LLM by setting the \emph{temperature} parameter to 0 for the entire experiment to produce deterministic outputs.\looseness=-1

\subsection{Metrics}\label{sec:metrics}
We employ three metrics to evaluate the explanations. Firstly, we use a \textit{root cause identification} metric which is marked as $1$ if the generated explanation contains the ground truth root cause of the failure/behaviour, and $0$ otherwise. This metric represents the ability of the explanation system to correctly identify the root cause of a failure or decision.

However, in some explanations, erroneous information is included, potentially in addition to correct information about the root cause. Thus, we also employ the \textit{presence of incorrect facts} metric, which is $1$ whenever the explanation contains false information about the failure or task, and $0$ otherwise. We merge the above into a combined \textit{explanation accuracy} metric, which is $1$ only if \textit{root cause identification} is $1$ and \textit{presence of incorrect facts} is $0$, and is $0$ otherwise.

In addition, we report the runtime required to compute the explanation, ensuring that all variants are evaluated on the same hardware. Specifically for HEXAR, we further measure the \textit{component explainer selection accuracy}, defined as the proportion of instances in which the explainer selector identifies the correct component explainer for a given scenario.

\subsection{Baselines}\label{sec:experiments_baselines}

\begin{figure} [t]
\centering
\includegraphics[width=0.47\textwidth]{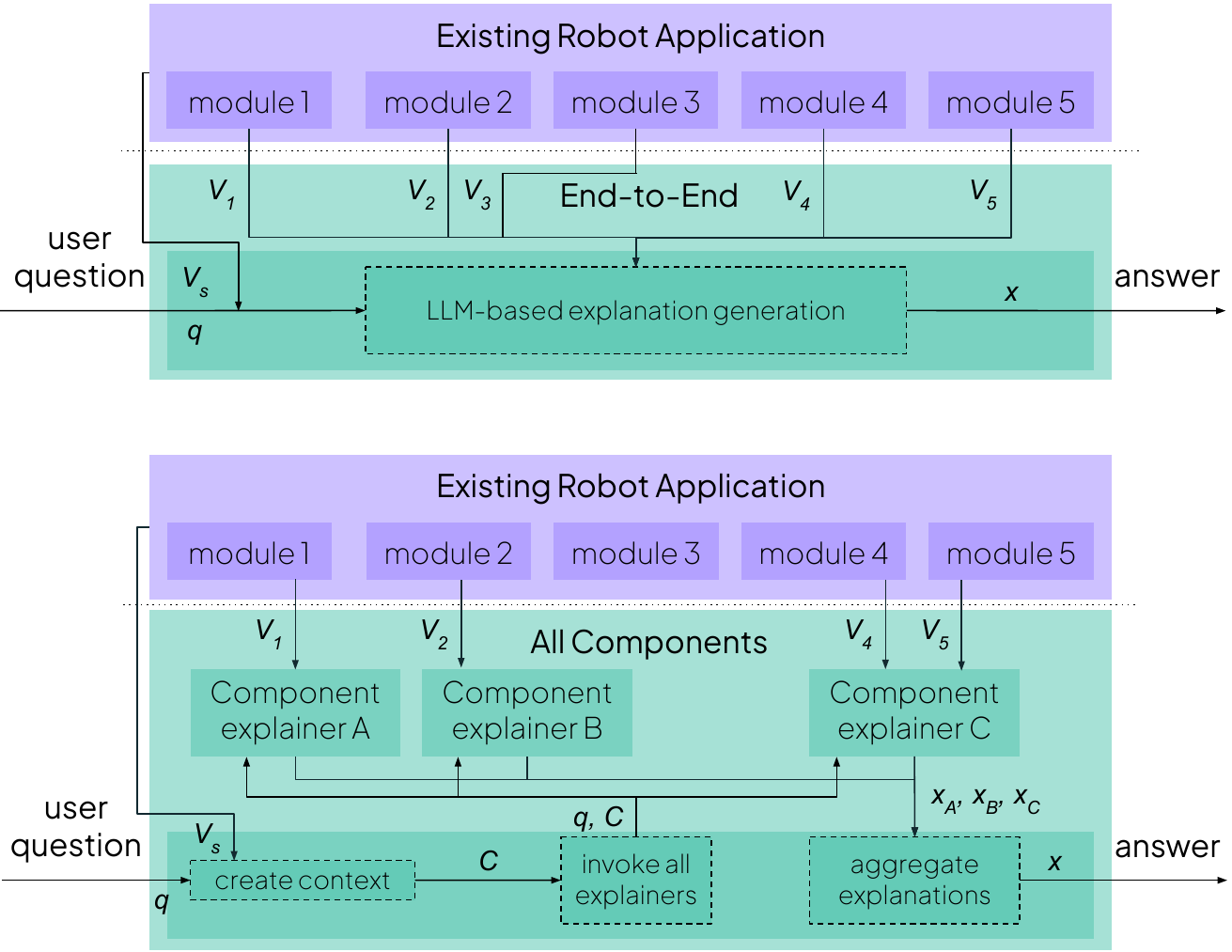}
\caption{Evaluated baselines: \textit{end-to-end} (top) and \textit{all components} (bottom).}
\label{fig:baseline-architecture_implementation}
\end{figure}

Representative of monolithic approaches based on LLMs, the \emph{end-to-end} baseline (Fig.~\ref{fig:baseline-architecture_implementation} top) uses a single LLM to parse all relevant logs and diagnostic information to produce an explanation in response to a query, similar to the approach in~\cite{sobrin2024explaining}. To ensure a fair comparison, this baseline is given all the same information used by \system, consisting of logs and parameter values. While this approach has access to all the same information, it lacks the specialised explainability of the component explainers and the discriminative power of the explainer selector.

To differentiate between the effects of the component explainers and the explainer selector, we also implement an \emph{all components} baseline (Fig.~\ref{fig:baseline-architecture_implementation} bottom), which is a modification of \system that retains all the component explainers. However, rather than selectively choosing a single component explainer to provide an explanation as in \system, it triggers all component explainers and then aggregates their outputs into a single explanation using an LLM, prompted only to select the relevant information. In this way, we examine the specific effect of selecting a single component explainer to answer a query.

We expect \system to outperform the two baselines in all the metrics defined in Sec. \ref{sec:metrics}. We also expect the \textit{end-to-end} baseline to perform worse than \textit{all components}, since it does not benefit from the specialised component explainers.\looseness=-1

\begin{table*}[t]
\centering
\begin{tabular}{llll}
\hline
\textbf{\#} & \textbf{Category} & \textbf{Relevant Module} & \textbf{Issue/Failure} \\
\hline
1 & Agent Error & Planning & The robot's planner produces a plan with an invalid skill \\
\hline
2 & Agent Error & Planning & The robot's planner produces a plan with invalid parameter names and/or values \\
\hline
3 & Agent Error & Planning & The robot's planner produces a plan which does not fulfil the user's request \\
\hline
4 & Inability & Planning & The robot is instructed to perform a task which it is unable to complete \\
\hline
5 & Unforeseen Circumstances & Navigation & Static obstacles prevent the robot from reaching a desired location \\
\hline
6 & Inability & Navigation & The robot's joystick controller is enabled, overriding autonomous navigation \\
\hline
7 & Inability & Navigation & The robot is plugged into its charger, overriding autonomous navigation \\
\hline
8 & Sub-Optimal Behaviour & Navigation & The robot is badly localised in its map, negatively impacting navigation \\
\hline
9 & Sub-Optimal Behaviour & Navigation & Moving obstacles force the robot to replan its path during navigation execution \\
\hline
10 & Normal/Successful & Navigation & No errors, but the user still questions the robot's movement properties (e.g. speed) \\
\hline
11 & Unforeseen Circumstances & Ask human for help & The robot does not detect anybody that can assist it in completing its task \\
\hline
12 & Inability & Ask human for help & The robot detects someone, but they are too far away to ask for help \\
\hline
13 & Uncertainty & Ask human for help & The robot detects someone, but not long enough for a stable detection \\
\hline
14 & Unforeseen Circumstances & Ask human for help & The robot detects someone, but is unable to approach them for help due to obstacles \\
\hline
15 & Unforeseen Circumstances & Ask human for help & The robot asks someone to assist it, but they refuse \\
\hline
16 & Unforeseen Circumstances & Ask human for help & Someone agrees to help the robot, but does not confirm completion of their assistance \\
\hline
17 & Social Norm Violation & Ask human for help & The robot approaches someone poorly due to suboptimal navigation \\
\hline
18 & Social Norm Violation & Ask human for help & The robot approaches someone poorly due to high variance in the person's detection \\
\hline
19 & Agent Error & Text to speech & The robot's text-to-speech skill times out before its utterance is complete \\
\hline
20 & Normal/Successful & Pizza recommender & The robot explains its choice of pizza with reference to available ingredients \\
\hline
\end{tabular}
\caption{Overview of the 20 situations in the evaluation dataset. Each situation is classified by the scenario category as defined by Wachowiak et al.~\cite{wachowiak2024people} and by the relevant module (skill).}
\label{tab:dataset}
\end{table*}

\subsection{Test cases}\label{sec:dataset}
To obtain test cases for evaluation, we begin by identifying situations in the home assistant robot use case that might require an explanation. Following a prior user study~\cite{wachowiak2024people}, we have identified $7$ broad categories of such situations that apply to our use case. By mapping each applicable category to a relevant module, we obtain 20 specific scenarios, as shown in Table \ref{tab:dataset}. To force the particular failure/behaviour specified by these scenarios, we configured the robot by changing default parameters in modules, such as lowering the timeout in \emph{text to speech} module or the tolerance threshold for variance in person localisation. For other scenarios, we modified the environment at runtime, e.g., by placing obstacles when the robot is navigating. To reliably simulate task planning errors, we manually injected the incorrect plans, bypassing the task planner module. For each scenario, we define a ground truth, a minimal description of the root cause of the failure/sub-optimal behaviour.

For each scenario, we introduce three variations of the user task instruction, which leads to a total of 60 task-specific scenario instances. We then execute the scenario on the TIAGo robot in a studio apartment environment and record all the necessary information required to produce explanations, using the \emph{rosbag} format. The same 60 \emph{rosbag} executions are used for each of the three methods tested. \looseness=-1   

For each scenario, we introduce variability by providing three distinct explanation-seeking queries, formulated as possible user questions regarding the failure/behaviour of the robot. The queries differ in their level of specificity, from generic (e.g., \textit{``What happened?''}) to more specific queries that either include the task context (\textit{``Why didn't you bring it?''}) or an observed problem (\textit{``Why did it take you so long?''}). By having 3 variations in the query for each of the 60 task-specific scenario instances, we obtain 180 explanations each from \system, \textit{end-to-end} and \textit{all components} baselines, totalling to 540 samples.

We generate the explanations by running the explainability-add on concurrently with the \textit{rosbag} and then providing a user query. To generate explanations, we use an 11th Gen Intel® Core™ i5-11400H @ 2.70GHz × 12 with 16 GB of RAM, where the LLMs are executed in an NVIDIA GeForce RTX 3080 with 12GB of memory. 


\subsection{Annotation Procedure}\label{sec:annotation_procedure}

To compute the metrics described in Sec. \ref{sec:metrics}, the explanations were labelled by 3 human annotators (co-authors) working independently and with a blind, randomised table including the testcase description, task instruction, user question, ground truth and generated explanation. Before the annotation phase, the 3 annotators discussed the metrics to have a common criterion in potentially ambiguous or corner cases. The final disagreement rate between the annotators was 0.93\% for the \textit{root cause identification} metric and 1.30\% for the \textit{presence of incorrect facts} metric. In Sec. \ref{sec:experiments_results}, we use a final value computed as the majority value across the annotators.\looseness=-1

\subsection{Results}
\label{sec:experiments_results}
\begin{figure}[t]
\centering
\includegraphics[width=0.49\textwidth]{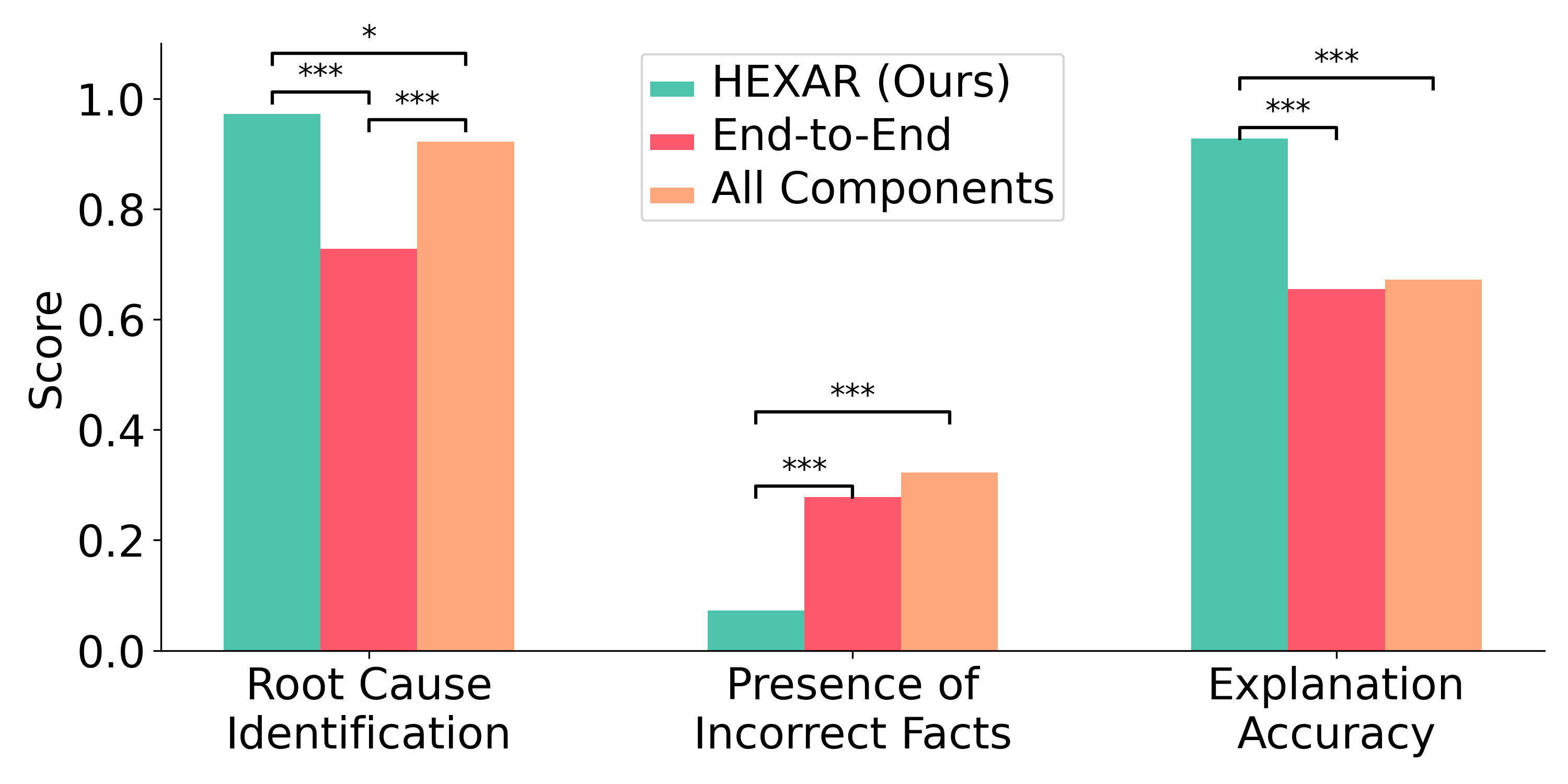}
\caption{Mean scores for each method, grouped by metric. For each pairwise difference, $*$ denotes $p<0.05$ and $***$ denotes $p<0.001$. Unlabelled pairs lack statistically significant difference.}
\label{fig:results}
\end{figure}

Across all scenarios (see Fig. \ref{fig:results}), HEXAR achieved the highest (best) mean \textit{root cause identification} rate of 97\% ($\sigma^2 = 0.16$), compared to the 73\% ($\sigma^2 = 0.45$) obtained by the end-to-end baseline and the 92\% ($\sigma^2 = 0.27$) obtained by the all components baseline. For the \textit{presence of incorrect facts} metric, HEXAR achieves the lowest (best) score at 7\% ($\sigma^2 = 0.26$), followed by the end-to-end baseline at 28\% ($\sigma^2 = 0.45$) and the all components baseline at 32\% ($\sigma^2 = 0.47$). For the combined \textit{explanation accuracy} metric, HEXAR achieves the highest (best) score at 93\% ($\sigma^2 = 0.26$), followed by the all components baseline at 67\% ($\sigma^2 = 0.47$) and the end-to-end baseline at 66\% ($\sigma^2 = 0.48$). We perform Cochran's Q test ($df = 2$) for the \textit{root cause identification} ($Q = 60.04, 
p < 0.001 $), \textit{presence of incorrect facts} ($Q = 45.50, 
p < 0.001$), and \textit{explanation accuracy} ($Q = 50.85, 
p < 0.001$) metrics, followed by pairwise McNemar tests with Holm correction to determine the statistical significance of results.

\begin{figure}[t]
\centering
\includegraphics[width=0.49\textwidth]{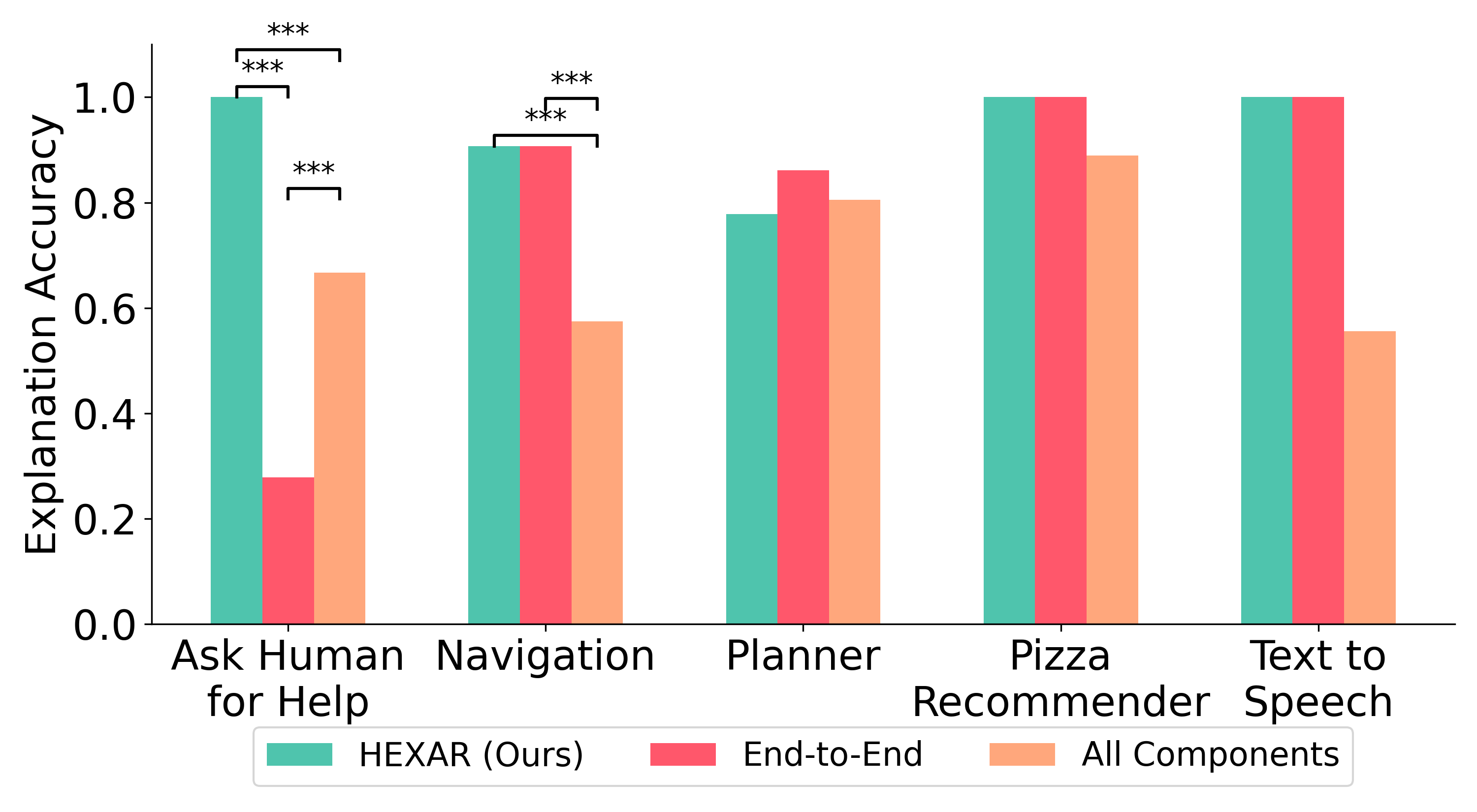}
\caption{Mean \textit{explanation accuracy} for each method, grouped by relevant robot module. All pairwise differences are not statistically significant, except those labelled $***$, for which $p<0.001$.}
\label{fig:results_by_component}
\end{figure}

Considering only the \textit{explanation accuracy} metric, we also compare performance across the five robot modules discussed in Sec. \ref{sec:implementation}. The same statistical tests are conducted, and results are presented in Fig. \ref{fig:results_by_component}.

Additionally, we note that across the 180 explanations, HEXAR correctly selects the appropriate component explainer in 179 cases, resulting in a \textit{component explainer selection accuracy} of 99.44\%.\looseness=-1

Finally, across all scenarios, we report a mean runtime of 1.73 seconds ($\sigma^2 = 1.66$) for HEXAR, compared to 7.86 seconds ($\sigma^2 = 2.04$) for the end-to-end baseline and 10.05 seconds ($\sigma^2 = 1.62$) for the all components baseline.

\section{Discussion}
\label{sec:discussion}
We now use the results described in the previous section to compare and contrast \system against two baselines. It is worth noting that these results are obtained from a single use case and implementation, and other \system implementations and use case particularities will necessarily alter the explanation generation performance. However, we assume that many of the patterns and considerations from our evaluation will continue to be valid, since they are related to the way information flows are structured and not particular to specific requirements, constraints or technologies.

Results indicate that the \textit{end-to-end} baseline has a significantly lower \textit{root cause identification} value than the other two versions, which we attribute to using ``raw'' information from the application in a centralised LLM instead of processing it through specialised modules that are better at identifying the root causes for particular modules. The \textit{all-components} baseline provides a lower value compared to \system, although it is somewhat higher than the \textit{end-to-end} baseline. This indicates that although the relevant component explainer response is necessary to find the root cause, irrelevant explanations from other components can add noise, which may obscure the root cause.

Considering the \textit{presence of incorrect facts} metric, the \textit{all components} baseline performs the worst, with similar values to the \textit{end-to-end} baseline. These results suggest that any extra irrelevant context, either first processed and then aggregated or directly aggregated, usually injects explanations with incorrect information, regardless of whether the explanation also points to the correct root cause. For example, in the scenario 7 from Table \ref{tab:dataset}, \system correctly produced ``I couldn't navigate because the robot is charging, which disables autonomous navigation.'', while the \textit{all components} baseline added a secondary, wrong statement about the high-level plan which was incorrect: ``The reason I didn't bring it [...] autonomous navigation was disabled while I was charging. Additionally, my plan was incorrect as I attempted to go directly to the living room without first navigating to the kitchen to pick up the coffee.''

Regarding the consolidated \textit{explanation accuracy} metric, it can be negatively affected either because of a missing root cause or by including incorrect information. Results indicate that the metric is higher for \system compared to the two baselines. These baselines perform relatively similarly, but for the \textit{all components} baseline, the \textit{presence of incorrect facts} is mainly responsible for the drop in accuracy, while in the case of the \textit{end-to-end} baseline, it is the \textit{root cause identification} that is the main contributor.

With respect to the runtime, results confirm that first selecting the relevant component explainer and executing only that dedicated module is more efficient than executing every single component explainer and then aggregating the information (\textit{all components} baseline) or having an LLM with a very extended context (\textit{end-to-end} baseline).

When comparing the five robot modules (Fig.~\ref{fig:results_by_component}), no baseline consistently outperforms \system across any category with any statistical significance. The \textit{end-to-end} baseline yields competitive performance in tasks where component explainers are LLM-based (i.e., navigation, planning) or where the explanations are relatively simple and repetitive (i.e., pizza recommender, text to speech). In contrast, this baseline exhibits difficulties in explaining more complex skills (ask human for help), where a tailored causal model provides superior identification of the actual root causes. For the \textit{all components} baseline, the poor performance in most categories other than planning—with the exception of the pizza recommender—is primarily attributable to the planner component introducing non-existent issues in the high-level plan, which are then incorporated into the final explanations alongside the correct root causes.

Finally, when analysing specifically the \system performance, we corroborate that the high \textit{component explainer selection accuracy} has been key to achieving the positive results. Only in one experiment was the wrong component explainer selected. In scenario 3 (Table \ref{tab:dataset}), where the planner incorrectly misses a step where the robot navigates to the living room in order to deliver a book, the system was asked ``Why didn't you go to the living room?''. The system responded with ``I do not have enough information to answer this question.'',  wrongly selecting the navigation component explainer instead of the planning component explainer.

\section{Limitations and Future Work}
\label{sec:future}

There are several promising directions for extending this work. In the current implementation, inter-dependence between component explainers was relatively limited. Future research could explore more complex instances of \system, designed to resemble robotic applications with additional modules, deeper inter-dependencies, greater structural depth, and more complex selection algorithms. Such extensions might involve assigning multiple component explainers to more complex modules, enabling explainers to invoke one another when appropriate, and systematically comparing the behaviour of different LLM models within this framework.

In this work, we considered only textual (i.e., natural language) explanations. However, many techniques produce explanations in other modalities, and future implementations should attempt to integrate these approaches into \system. For example, to explain human detection error or uncertainty, visual explanation components such as bounding boxes can be coherently combined with a textual component to explain how a lower-level module leads to task failure.\looseness=-1

The evaluation procedure and metrics used in this work allowed us to objectively determine which systems produced accurate explanations containing root causes and excluding incorrect information, thus sufficiently answering our research question. In the future, user studies can be performed to further evaluate subjective attributes and the effects of explanations on users.

Finally, beyond providing explanations, \system could be extended to support replanning, reactive failure recovery and prevention, as well as the generation of corrective action recommendations for the user.

\section{Conclusions}
\label{sec:conclusions}
We have presented \system, a hierarchical explainability architecture for robots that addresses the challenge of explaining modular robotic systems through specialised component explainers orchestrated by a selector. Our framework provides a plug-in approach that leverages existing robot interfaces.\looseness=-1


We provide an example implementation of the framework for a real robot performing assistive tasks in a home environment. We provide evidence that \system significantly outperforms two baselines: an end-to-end approach using a single LLM and another approach that first invokes all component explainers and then aggregates their outputs. Based on 20 scenarios where situations requiring explanations are recreated, we validate that \system is more effective at identifying root causes and avoiding incorrect information while being more time-efficient, affirmatively answering our research question. Our approach not only enhances explanation accuracy but also facilitates the integration of heterogeneous explanation techniques tailored to various robot capabilities with different complexities and dependencies. As robotic systems continue to grow in complexity, hierarchical explainability architectures offer a promising path toward transparent autonomous robots.

\balance


\section*{Acknowledgements}

This work has been supported by Horizon Europe Marie Skłodowska-Curie grant agreement No. 101072488 (TRAIL) and the Horizon Europe CoreSense project (grant 10107025).



\bibliographystyle{ieeetr}
\bibliography{bibliography}

\end{document}